\documentclass{egpubl}
\usepackage{eg2018}

\ConferenceSubmission   
 \electronicVersion 

\ifpdf \usepackage[pdftex]{graphicx} \pdfcompresslevel=9
\else \usepackage[dvips]{graphicx} \fi

\PrintedOrElectronic

\usepackage{t1enc,dfadobe}

\usepackage{egweblnk}
\usepackage{cite}

\usepackage{amsmath}
\usepackage{amssymb}
\usepackage{subfigure}
\usepackage{rotating}
\usepackage{multirow}
\usepackage{color}
\usepackage{graphicx}
\usepackage{times}
\usepackage{balance}
\usepackage{array}
\usepackage{graphicx}
\usepackage{epstopdf}
\usepackage{flushend}
\usepackage{amsmath, amssymb, amsfonts, amsopn}

\def\eg{\emph{e.g.}}

\begin{document}

\title[Learning Semantic Abstraction of Shape via 3D Region of Interest]%
      {Learning Semantic Abstraction of Shape via 3D Region of Interest}



\author[H. Fang and X. Wang et al.]{Haiyue Fang\thanks{Joint first author}, Xiaogang Wang\footnotemark[1],  Zheyuan Cai, Yahao Shi, Xun Sun, Shilin Wu, Bin Zhou\thanks{Corresponding author:zhoubin@buaa.edu.cn}\\
State Key Laboratory of Virtual Reality Technology and Systems, School of Computer Science and Engineering, Beihang University}



\maketitle
\begin{abstract}
   In this paper, we focus on the two tasks of 3D shape abstraction and semantic analysis. This is in contrast to current methods, which focus solely on either 3D shape abstraction or semantic analysis. In addition, previous methods have had difficulty producing instance-level semantic results, which has limited their application. We present a novel method for the joint estimation of a 3D shape abstraction and semantic analysis. Our approach first generates a number of 3D semantic candidate regions for a 3D shape; we then employ these candidates to directly predict the semantic categories and refine the parameters of the candidate regions simultaneously using a deep convolutional neural network. Finally, we design an algorithm to fuse the predicted results and obtain the final semantic abstraction, which is shown to be an improvement over a standard non maximum suppression. Experimental results demonstrate that our approach can produce state-of-the-art results. Moreover, we also find that our results can be easily applied to instance-level semantic part segmentation and shape matching.


\printccsdesc   
\end{abstract}  

\section{Introduction}
\begin{figure}
\begin{center}
   \includegraphics[width=\linewidth]{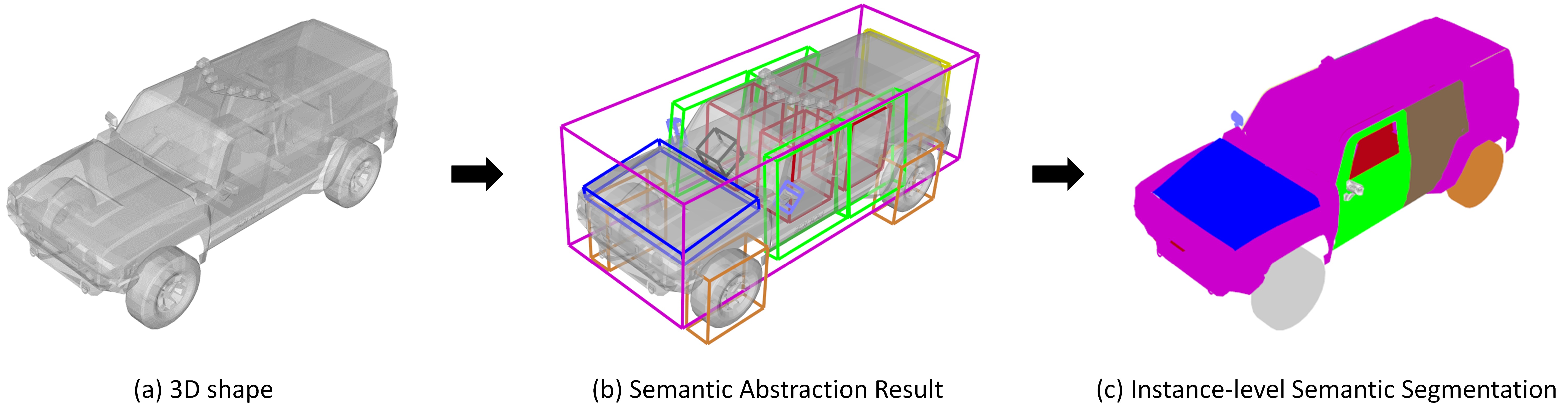}
\end{center}
   \caption{3D shape abstraction and semantic analysis.}
\label{fig:teaser}
\end{figure}

 Current methods of 3D shape analysis focus solely on either abstraction or semantic analysis. In addition, it is difficult to obtain instance-level semantic results from current 3D shape semantic analysis methods, which facilitates part-based
shape synthesis and modeling~\cite{zhu2018scores, xu2016data}. 
In this paper, we focus on the two tasks of 3D shape abstraction and semantic analysis, as shown in Figure~\ref{fig:teaser},
We introduce a novel method for the estimation of shape semantic abstraction. 
Inspired by Faster-RCNN~\cite{ren2015faster},
we use semantic abstraction candidates as building  blocks for the 3D shape semantic abstraction.
The choice is motivated by the success of proposal based image detection.
Our semantic abstraction method operates in three stages.
In the first stage, we extract semantic abstraction candidates for each 3D shape according to the distribution of semantic parts.
For semantic abstraction positions, a Gaussian Mixture Model (GMM) is trained to obtain a position probability distribution function.
For different semantic parts, the scale varies greatly,
so we define $N$ different scale primitives for each specific semantic category by clustering.
The semantic abstraction template is obtained by combining $N$ basic scale primitives with each candidate position.
Finally, we extract semantic abstraction candidates by placing the set of abstraction template into the 3D shape.
This method can greatly reduce time consumption compared to an exhaustive search.

In the second stage, a classifier  scores the semantic abstraction candidates and a regressor refines the parameters of abstract primitives.
The classifier and the regressor share the convolutional feature layer and are trained jointly.
In the third stage, the final semantic abstraction estimation is obtained by integrating over neighboring semantic abstraction results; that is shown to improve over a standard non-maximum suppression.

We demonstrate the performance of our algorithm on our benchmark dataset.
The 3D shapes in our dataset are collected from ShapeNet~\cite{ShapeNet2015} and 3D warehouse~\cite{Tri3Dwarehouse}. We manually annotate each 3D shape
in our dataset using our interactive annotation tool. The annotation tool is elaborated in the supplementary material.

In addition to the main contribution of providing a method to perform
3D shape semantic abstraction, our two secondary contributions are:
1),a novel abstraction template is learned using a semantic part distribution,
and we then generate semantic abstraction candidates by placing the semantic abstraction template into a 3D shape.
This helps to address the challenging problem of an exhaustive search resulting in excessive time consumption (Section 3.1);
2), we provide a benchmark dataset which includes 4 categories and 5022 shapes in total (Section 4.1).
This immediately benefits many tasks, such as shape modeling, shape deformation, etc.
Moreover, we find that our results can be easily applied to instance-level semantic part segmentation and shape matching.

This paper is organized as follows: Section 2 reviews the related literature,
Section 3.1 introduces the algorithm used to generate semantic abstraction candidates for a 3D shape
and Section 3.2 presents the network architecture. Section 3.3 introduces the semantic abstraction
integration algorithm.
Section 4 presents experimental results and shows comparisons with the state of the art. 
We conclude and discuss ideas for future work in Section 5.


\section{Related Work}

Our work is related to: 3D shape abstraction and shape semantic analysis. We will review the most relevant work on these topics in the remainder of this section.


\subsection{Shape abstraction}
3D shape abstraction is a fundamental problem. In recent years, many shape abstraction techniques have been developed.
An intuitive shape abstraction technique is shape simplification, which involves merging adjacent triangles to obtain the shape abstraction~\cite{Garl_1997}.
For mainfold models, such methods can achieve exciting results; however, for man-made shapes, the most commonly seen data form in modern 3D shape repositories (\eg, Trimble 3D Warehouse~\cite{Tri3Dwarehouse} and ShapeNet~\cite{ShapeNet2015}), this approach does not achieve satisfactory results and may lead to loss of shape structure.
In addition to shape simplification, another technique for shape abstraction is contour description, which uses lines to express the model shape contour
~\cite{Hild_05,Lai2007,gehre_16,Mehra2009,Goes2011,Gori_17,tsai2017user}.
Goes et al.~\cite{Goes2011} propose the concept of an exoskeleton as a new abstraction of arbitrary shapes that succinctly conveys both the perceptual and the geometric structure of a 3D model. They extract exoskeletons via a principled framework that combines segmentation and shape approximation.
However, such approaches only work with manifold models.
Mehra et al.~\cite{Mehra2009} introduce a method for abstracting 3D man-made models using characteristic curves or contours as building blocks for the abstraction. They use a two-step procedure that first approximates the input model using a manifold, closed envelope surface and then extracts from it a hierarchical abstraction curve network along with suitable normal information.
Their approach can handle any type of 3D shape, but abstracts only the appearance of the 3D shape.
Meanwhile, for an unsupervised shape abstraction method, it is difficult to guarantee the quality of the shape abstraction.
For this reason, Kratt et al.~\cite{Kratt2018} introduce user interaction information in the process of model abstraction; that is, shape abstraction is conducted under the guidance of user requirements.

Our technique is most related to the primitives based technique. Such methods assume that the shape abstraction preserves the original function and structure of the model~\cite{Coab2012,Coco2014,ZFDOT10,Laga_2017}.
Yumer et al.~\cite{Coco2014} defined three types of surface, namely plane, surface and sphere. Based on the contour expression, the original contour lines were fitted into three types of surface and then abstracted.
Heng et al.~\cite{ZFDOT10} defined four basic primitives: ball, cube, cylinder and generalized cylinder. 
There are editing operations corresponding for each primitive. The primitive is used to replace the part according to the geometric features of the shape components. Laga et al.[15] used cubes and cylinders for 3D shape abstraction.
Yumer et al.~\cite{Coab2012} presented a co-abstraction method that takes as input a collection of 3D shapes, and produces a mutually consistent and individually identity-preserving abstraction of each shape.

At the same time, with the rapid development of neural networks, there are also many shape abstraction methods based on neural networks~\cite{Song2017,abst_17}.
Song et al.~\cite{Song2017} proposed a multi-view based shape abstraction method. Firstly,they render a 3D shape into images under different perspectives, and detect the potential parts (as boxes) in the images using Faster RCNN~\cite{ren2015faster}. Finally, they select the vertices for each part category, and generate the 3D bounding boxes.
However, faces hidden from the surface of the 3D shape are invisible to any view, and thus cannot be abstracted.
Tulsiani et al.~\cite{abst_17} proposed an unsupervised neural network approach that abstracts the input 3D shape, which consists of multiple directed bounding boxes.
However,for complex 3D shapes with overlap, unsupervised methods only require that the abstract primitives can cover the 3D shape as much as possible, while the primitive is as small as possible. Hence such approaches are naturally unable to obtain meaningful abstraction for 3D shapes with complex intersections.

\begin{figure}
\begin{center}
   \includegraphics[width=\linewidth]{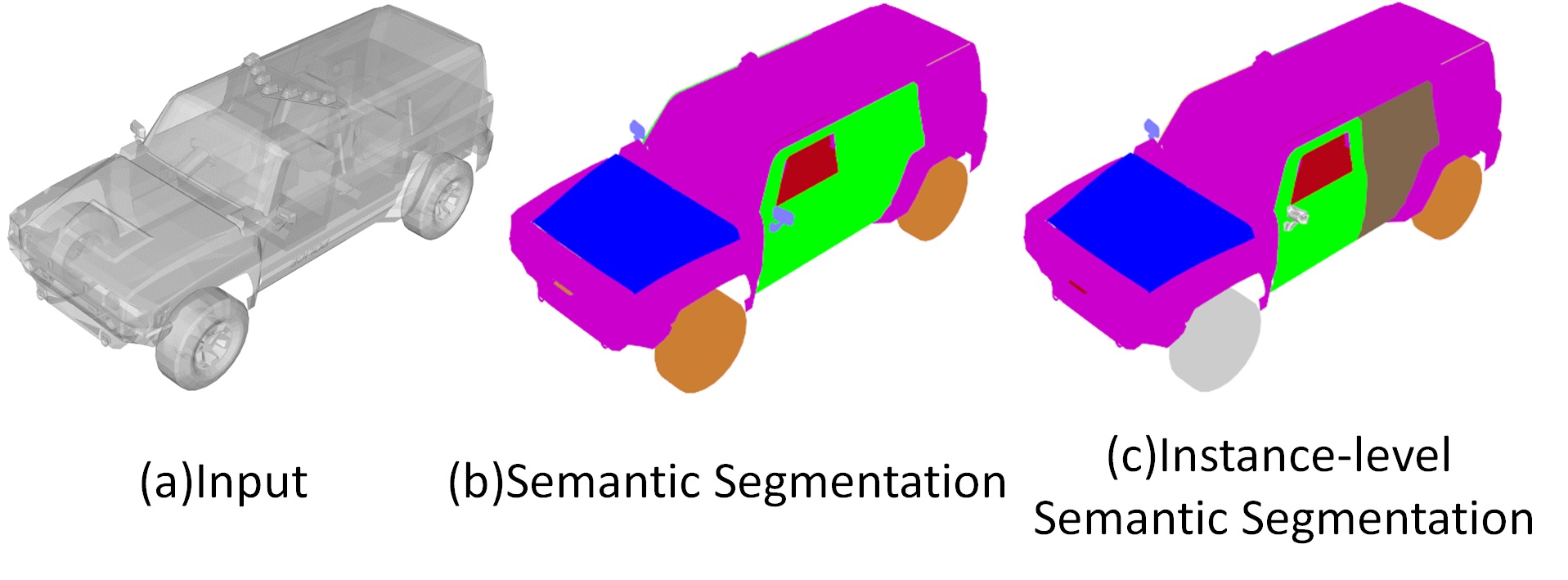}
\end{center}
   \caption{Semantic segmentation vs. instance-level semantic segmentation.}
\label{fig:instance_vs_seg}
\end{figure}

\begin{figure*}
\begin{center}
   \includegraphics[width=\linewidth]{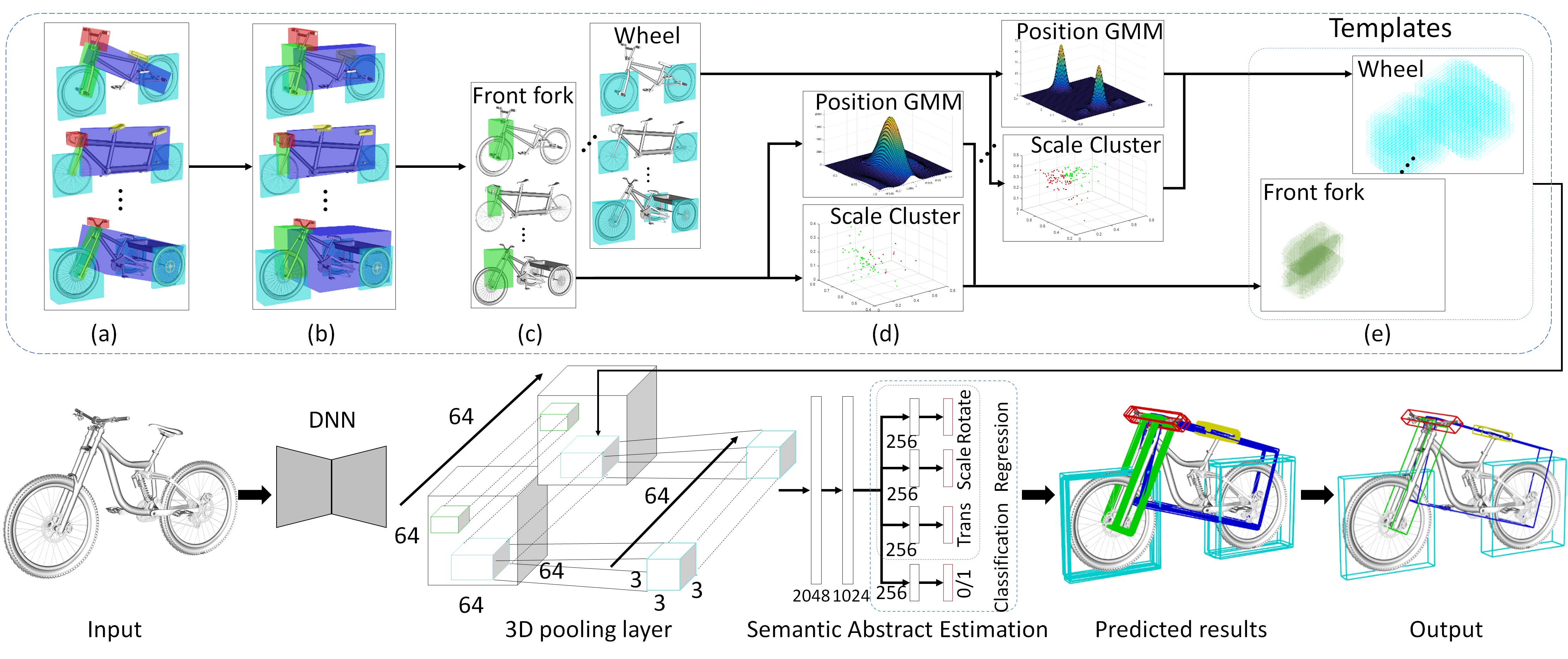}
\end{center}
   \caption{Overview of our semantic abstraction framework. 
   We first learn the semantic abstraction template using statistic cues (top) 
   and generate semantic abstraction candidates by placing the abstraction template into the input 3D shape. 
   These semantic abstraction candidates are then scored by a classification branch and regressed using a regressor. Finally, the semantic abstraction estimation is obtained by integrating over neighboring predicted results.}
\label{fig:overview}
\end{figure*}
\subsection{Semantic analysis of shape}
Many methods have been proposed for shape segmentation and labeling.
Early studies~\cite{Katz_SG03,Huang_EG09,Shapira_IJCV10,Zhang_TOG12,Au_TVCG12} aimed to utilize hand-crafted geometry features for mesh segmentation and labelling,
such as curvature~\cite{Gal_TOG06},
PCA~\cite{Kalogerakis_SG10},
shape diameter function~\cite{Shapira_IJCV10},
distance from medial surface~\cite{Liu_EG09},
average geodesic distance~\cite{Hilaga_SG01},
shape context~\cite{Belongie_PAMI02},
spin image~\cite{Johnson_PAMI99}, etc.

Recently, deep learning-based 3D shape semantic analysis has made great progress~\cite{Kalogerakis_SG10,Xie_SGP14,Guo_TOG15,Yi_CVPR17}.
Guo et al.~\cite{Guo_TOG15} learned a compact representation
of triangles for 3D mesh labeling
by non-linearly combining and hierarchically compressing various geometry features with the deep CNNs.
Li et al.~\cite{Yi_CVPR17} proposed a method, named SyncSpecCNN,
to label the semantic parts of 3D mesh.
The SyncSpecCNN trained vertex functions using CNNS,
and conducted spectral analysis to enable kernel weight sharing
by using localized information of mesh graph.

Recently, Kalogerakis et al.~\cite{Kalogerakis_CVPR17} proposed a deep architecture for
segmenting and labeling semantic parts of 3D shape
by combining image-based multi-view Fully Convolutional Networks and surface-based CRFs.
The projection-based methods~\cite{Kalogerakis_CVPR17,Wang_SA13} are suitable for imperfect (\eg, incomplete, self-intersecting, and noisy) 3D shapes, while not considering the `within-without' structure of raw 3D CAD shapes.
Su et al.~\cite{Su_CVPR17} designed a novel type of neural network, named PointNet,
for directly segmenting and labeling the point clouds
with respect to the permutation invariance.
PointNet showed strong performance on par with or even better than the state of the art.
However, such approaches can only obtain semantic-level segmentation results, and cannot get the instance-level results: 
as shown in figure~\ref{fig:instance_vs_seg}(b), all wheels and doors are belong to the same semantic category, and cannot parse out separately. This kind of result also greatly limits its range of application.

\section{Approach}
\label{sec:approach}

\subsection{Generation of Semantic Abstraction Candidates}
\textbf{Semantic abstraction template learning.  }
To reduce the time consumption, we do not perform an exhaustive search for semantic abstraction candidates directly. 
We first learn a semantic abstraction template from the annotated data set, according to the distribution of semantic parts.

Because what we label is the Oriented Bounding Box (OBB) for the semantic parts of the 3D shape in our dataset, and it contains different rotations,
it is difficult to generate statistics directly. 
To simplify the difficulty of statistics, we first convert the annotated OBB to Axis Align Bounding Box (AABB), as shown in Figure~\ref{fig:overview}(b). For each semantic category of 3D shape, we find that it can be fitted by a Gaussian Mixture Model (GMM),
and we are able to obtain a position probability distribution function.
We then take a discrete sampling of the probability distribution function
to generate semantic abstraction candidate positions, as shown in Figure~\ref{fig:overview}(d).
We learned GMM parameters using the standard MATLAB toolbox, with the number of Gaussian functions being $K$.
For different semantic categories, K is completely different, which is difficult to be directly specified. In this paper, we use AP (Affinity Propagation Clustering Algorithm) algorithm to cluster the center position of the bounding box, which can automatically identify satisfactory cluster centers. The number of identified clusters is set as the number of Gaussian functions.

To reduce the time and space overhead, and not diminish sampling quality, the three-dimensional continuous space
is discretized as $100 \times 100 \times 100$; that is, the step size is set to 0.01.
Let $C'$ be the set of all possible 3D positions, while the probability density of each position is $p_{c}$.
We can obtain the semantic abstraction candidates location set $C$, in which $p_{c}$ is larger than a threshold $Th_{loc}$.
\begin{eqnarray}
\begin{aligned}
  C=\{c \in C'| p_{c}>Th_{loc} \}.
\label{eq:CRF}
\end{aligned}
\end{eqnarray}

\begin{figure}
\begin{center}
   \includegraphics[width=1\linewidth]{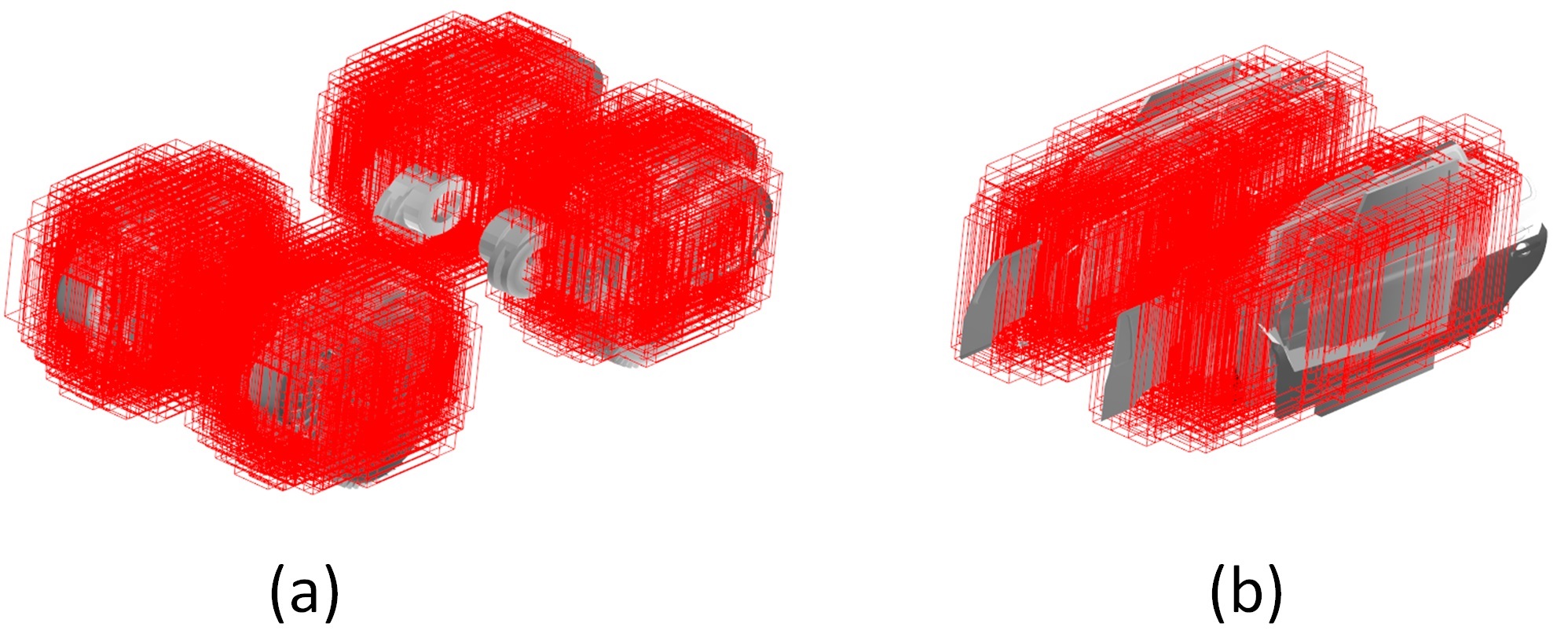}
\end{center}
   \caption{Semantic abstraction template (red boxes) for two semantic categories. (a) and (b) are the vehicle wheels and doors respectively.}
\label{fig:template}
\end{figure}

Meanwhile, for different semantic parts, the scale varies greatly. Therefore, we define $N$ different scale primitives for each semantic category by clustering. In our paper, $N$ is set to 20. 
Finally, the semantic abstraction template is obtained by placing $N$ basic scale primitives on each candidate position, i.e., $T_{c}=\{T_{c1}, T_{c2},..., T_{cN}\}$.
Figure~\ref{fig:template}(a) and (b) represent the semantic abstraction template (red boxes) for the vehicle wheels and doors.

\begin{figure}
\begin{center}
   \includegraphics[width=0.8\linewidth]{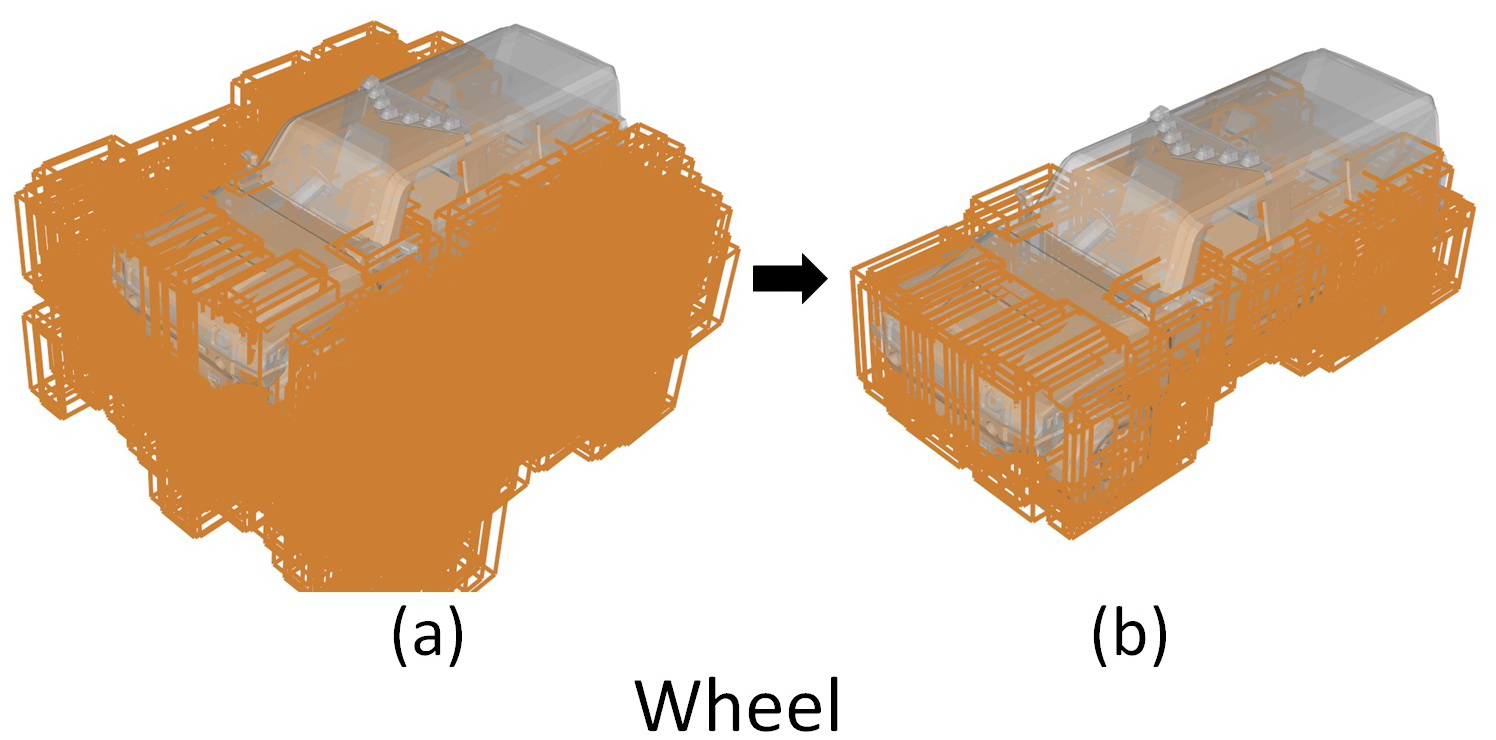}
\end{center}
   \caption{Semantic abstraction template shrinkage. (a) and (b) represent the semantic template and semantic abstraction candidates, respectively.}
\label{fig:shrink}
\end{figure}

\noindent
\textbf{Semantic abstraction candidates generation.}
If a semantic abstraction template is directly nested on a given 3D shape, a large amount of redundant space will be generated; that is, the template box cannot be compactly attached to the 3D shape.
As shown in Figure~\ref{fig:template}(a), the boxes of the vehicle wheels template are placed directly on the car and cannot be attached tightly to the car.
If the template is directly used as semantic abstraction candidate, the quality is poor. It also increases the difficulty of training,
especially the regression task.
Therefore, we effectively shrink each semantic template box to fit on the given 3D shape as a semantic abstraction candidate(see Figure~\ref{fig:shrink}(b)).

\subsection{SAE-Net Architecture}
\label{SAE-Net}

\noindent
\textbf{Procedure.}
Each input 3D shape is converted into a $64 \times 64 \times 64$ volumetric representation that is suitable for input to neural networks.
The volumetric representation is then fed to a U-net architecture, and the output is a $64 \times 64 \times 64 \times 64$ feature map.
Then, the features in the semantic abstraction candidate are converted into a small feature map with a fixed spatial extent 
and two fully connected layers, to jointly conduct semantic abstraction candidate classification and regression.

The encoder consists of 5 spatial convolution layers with numbers of channels $\{$16; 64; 128; 256; 2048 $\}$, and kernel sizes $\{$3; 3; 3; 3; 4$\}$, respectively. 
The decoder consists of 5 upconvolution layers with numbers of channels
$\{$256; 128; 64; 16; 64$\}$, and kernel sizes $\{$4; 3; 3; 3; 3$\}$, respectively. 
There are ReLU and batch normalization layers in between. 
In our architecture, these upconvolved feature maps are concatenated with the corresponding feature maps from the encoder part of the CNN. The output is a $64 \times 64 \times 64 \times 64$ feature maps.

\noindent
\textbf{3D RoI pooling layer.}
Similar to ~\cite{girshick15,ren2015faster}, the 3D RoI pooling layer uses max pooling to convert the features inside any semantic abstraction candidate into a small feature map with a fixed spatial extent of $H \times W \times L$, where $H ,W$ and $L$ are hyper-parameters of 3D RoI pooling layer. In this paper, the hyper-parameters are set to $3 \times 3 \times 3$.  3D RoI pooling works by dividing the $h \times w \times l$ abstraction candidate box into an $H \times W \times L$ grid and then max-pooling the values in each sub-window into the corresponding output grid cell.

\noindent
\textbf{Training sampling.}
We label all semantic abstraction candidates, as positive if their 3D IoU scores with ground truth are larger than 0.5, and negative if their IoU are smaller than 0.3. For the classification and regression loss, we use 64 ion candidates per batch, with 8 candidates from 8 different 3D shapes. Half the candidates are positive, 
and the remaining ones are negative. In order to balance the training samples in each semantic category, we select the training samples uniformly in each  semantic category, for both positive and negative candidates.

\begin{figure}
\begin{center}
   \includegraphics[width=1\linewidth]{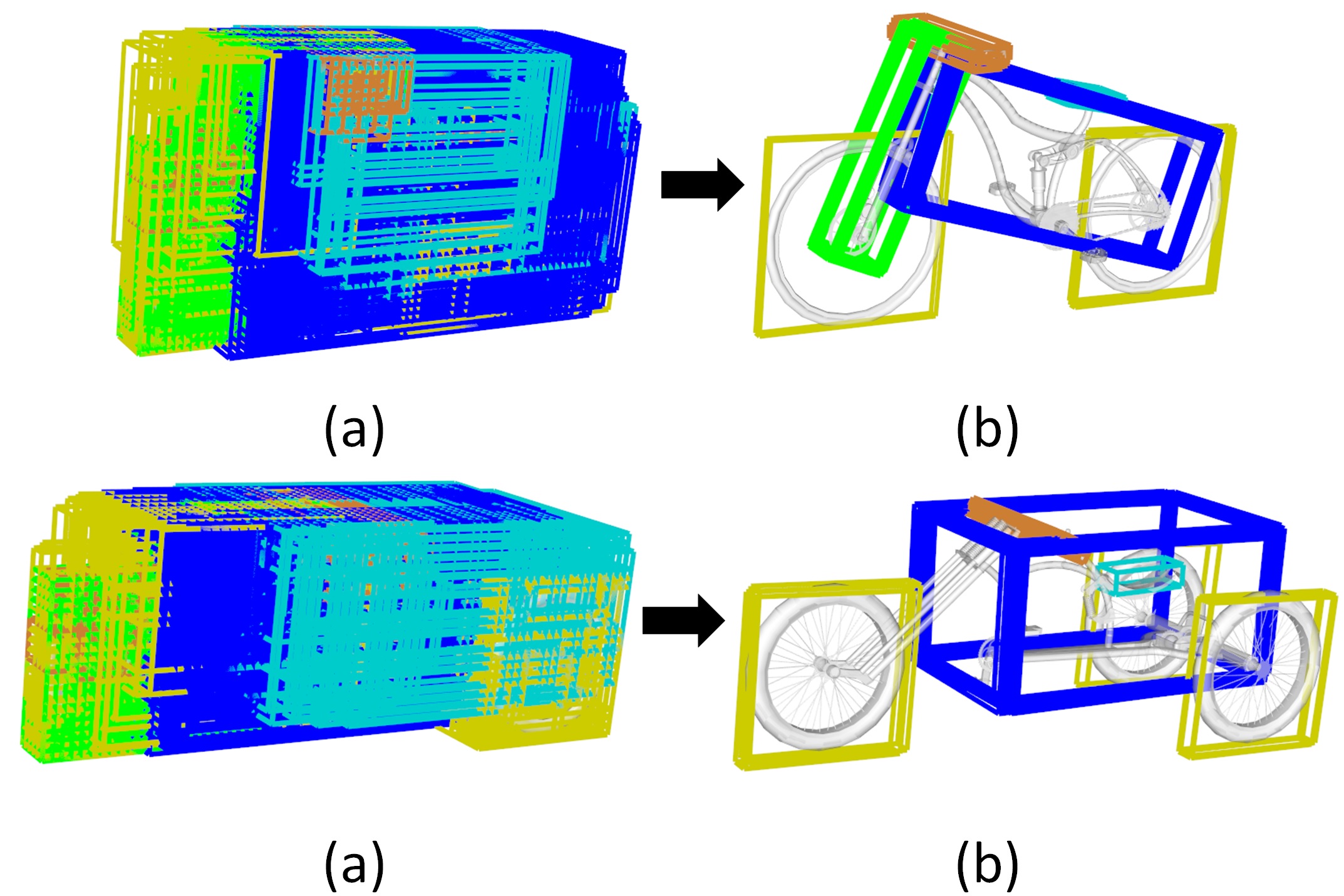}
\end{center}
   \caption{Two examples of bicycle semantic abstraction predictions.}
\label{fig:predict}
\end{figure}

\noindent
\textbf{Multi-task loss.}
We represent each 3D semantic abstraction candidate by its center [$c_x, c_y, c_z$], scale size [$s_1, s_2, s_3$] in three major directions and rotation angle [$a_x, a_y, a_z$]. To train the 3D semantic abstraction regressor, we divide the task into three sub-tasks: translation, scale and rotation. Therefore, we will predict the difference of centers, sizes and rotation angles between a semantic abstraction candidate and its ground truth annotation. For each positive semantic abstraction candidate and its corresponding ground truth, we represent the offset of semantic abstraction centers by their difference $c$=[$\bigtriangleup c_x, \bigtriangleup c_y, \bigtriangleup c_z$ ]. For the size and rotation angle difference, we first search the closest matching of major directions between the abstraction candidate and its corresponding ground truth and calculate the offset of scale size $s$=[$\bigtriangleup s_1, \bigtriangleup s_2, \bigtriangleup s_3$ ] and rotation angle $a$=[$\bigtriangleup a_x, \bigtriangleup  a_y, \bigtriangleup  a_z$ ].
Since each candidate abstraction automatically carries semantic information, our classification task is binary, based on the current semantic abstraction is good or not. 
For each semantic abstraction candidate, our multi-task loss function is defined as:

\begin{eqnarray}
\begin{aligned}
  L(p,p*,c,&c*,s,s*,a,a*)=\\
  &L_{cls}(p,p*)+\lambda p* L_{reg}(c,c*,s,s*,a,a*).
\label{eq:CRF}
\end{aligned}
\end{eqnarray}
where $L_{cls}$ is the binary confidence score, and $L_{reg}$ is the semantic abstraction regression. 
p is the predicted probability of this abstraction candidate being a reasonable semantic abstraction 
and p* is the ground truth (1 if the semantic abstraction candidate is positive, and 0 if the semantic abstraction candidate is negative).
 $L_{cls}$ is the log loss over two categories (is $vs.$ is not a reasonable semantic abstraction).
The semantic abstraction regression $L_{reg}$ is defined as:

\begin{eqnarray}
\begin{aligned}
  L_{reg}(c,c*,&s,s*,a,a*)=\\
  &L_{cent}( c,c*)+L_{rota}( s,s*)+L_{scale}(a,a*).
\label{eq:CRF}
\end{aligned}
\end{eqnarray}
where $L_{cent}$, $L_{rota}$ and $L_{scale}$ are all smooth L1 loss, a robust version of the L2 loss, which is less sensitive to outliers~\cite{ren2015faster}.
The semantic abstraction candidates and the prediction abstraction results using the network
are shown in Figure~\ref{fig:predict}(a) and (b), respectively.

\subsection{Composite Inference}
\label{Composite}

Our Semantic Abstraction Estimation Network (SAE-Net) outputs a set of refined semantic abstraction candidates with associated confidence scores. 
Multiple candidates cover each semantic part in the 3D shape.
One way to address this is to use Non-Maximum Suppression (NMS) to obtain the candidate with the highest scoring candidates as the estimated semantic abstraction.
However, we propose to aggregate the semantic abstraction candidates that are close in terms of semantic categories and 3D position.
 We refer to this post processing stage as Semantic Abstraction Integration (SAI).

\begin{figure}
\begin{center}
   \includegraphics[width=0.8\linewidth]{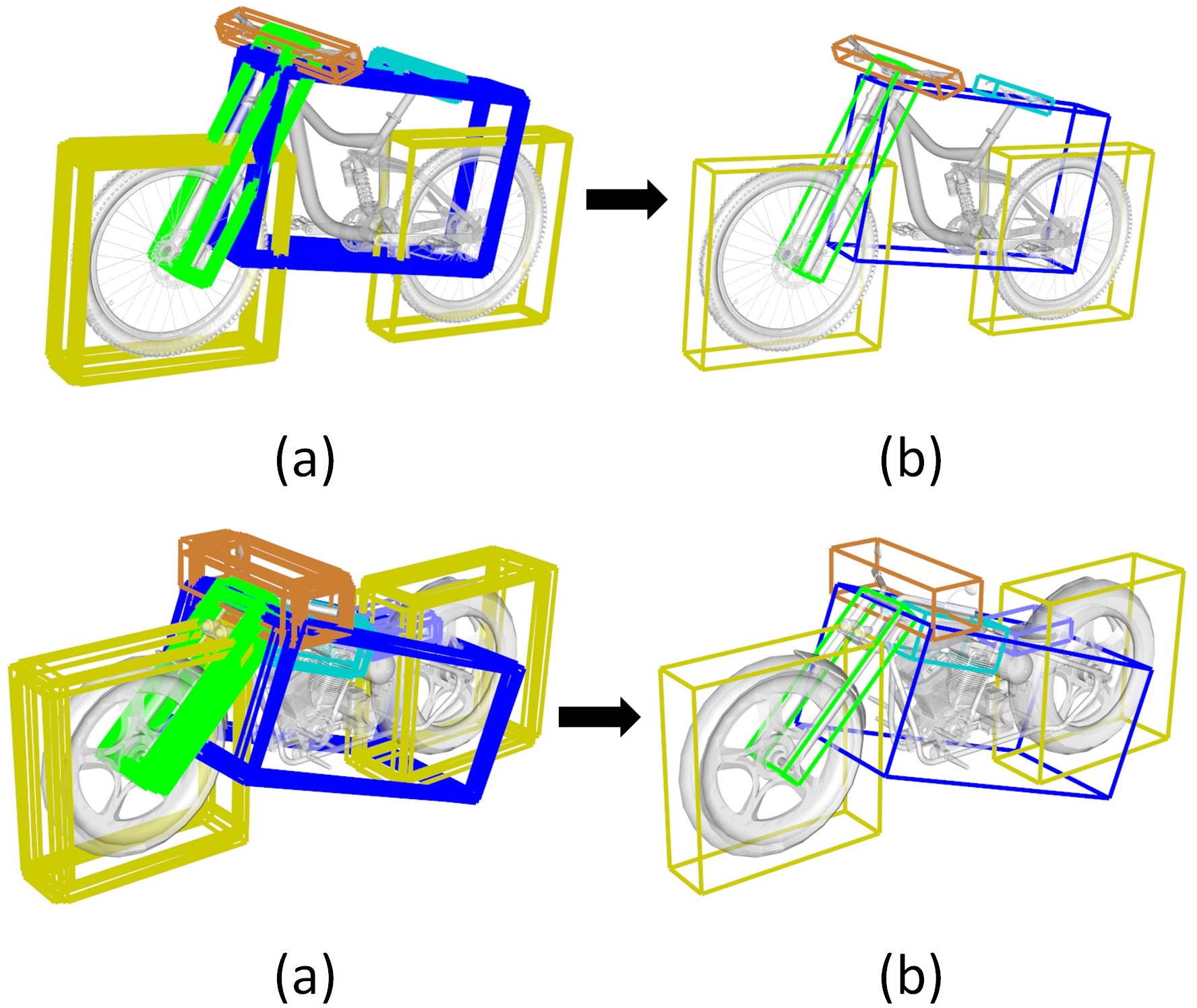}
\end{center}
   \caption{Two examples for bicycle semantic abstraction integration(SAI).}
\label{fig:post_process}
\end{figure}

For each semantic category, we first take the top scoring semantic abstraction candidate in the group and obtain all the semantic abstraction candidates that overlap sufficiently with this top scoring candidate.
We repeat this step with the remaining candidates and their top scoring elements until no candidates are left.
Therefore, the spatial overlap and semantics of the resulting groups are consistent.
Finally, we obtain semantic abstraction $a$ in candidates group $G$ weighted by their confidence score:

\begin{eqnarray}
\begin{aligned}
  a=\frac{1}{S} \sum_{g \in G} s(g) * g.
\label{eq:CRF}
\end{aligned}
\end{eqnarray}
in which $S$ is the sum of the individual confidence scores in group $G$, $i.e.$, $S=\sum_{g \in G}s(g)$.
Figure~\ref{fig:post_process} shows two examples for composite inference; (a) and (b) represent the prediction abstraction results and the final semantic abstraction, respectively.

\section{Experiments}
\begin{figure*}
\begin{center}
   \includegraphics[width=\linewidth]{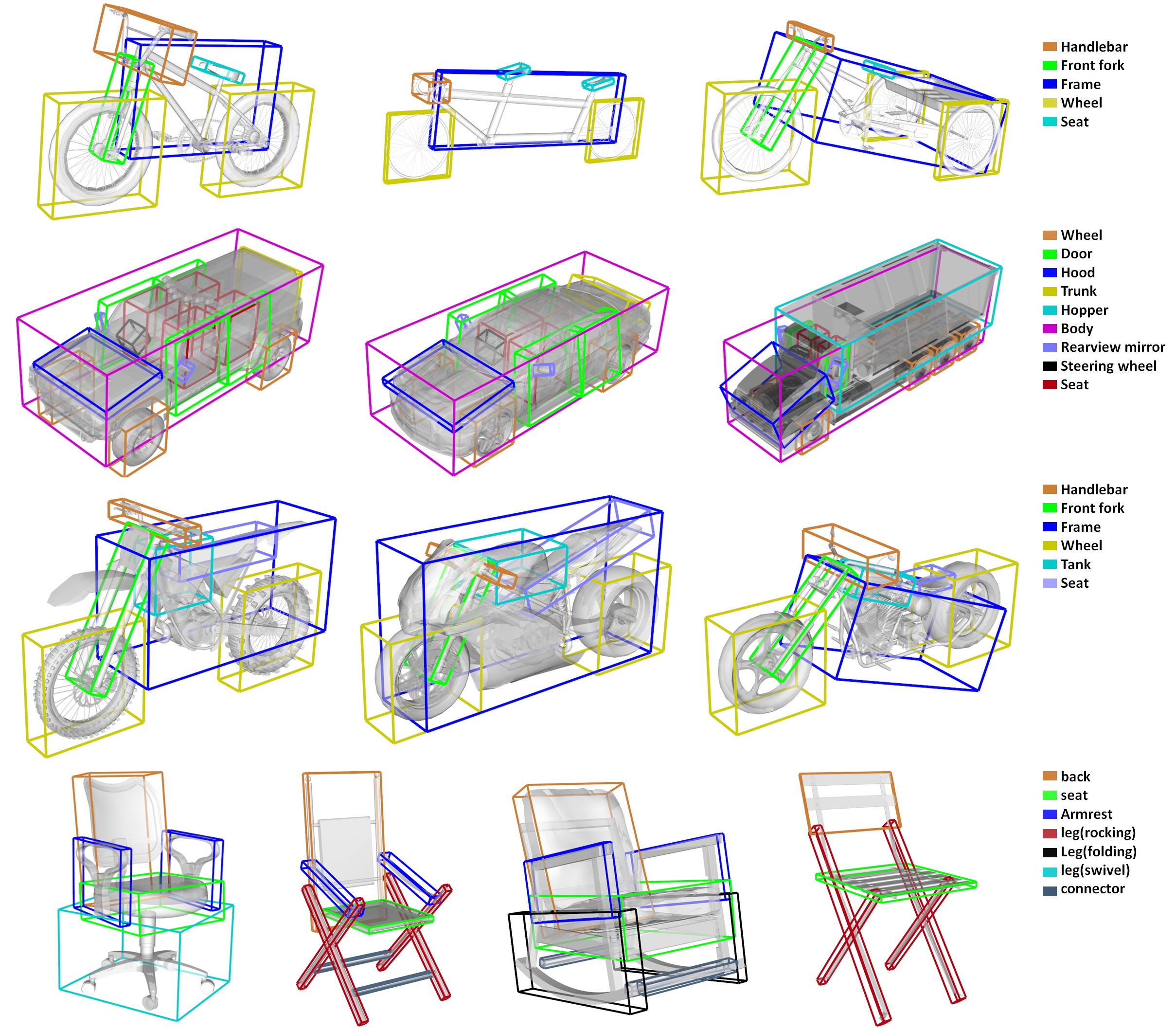}
\end{center}
   \caption{Overview of our benchmark dataset.}
\label{fig:dataset}
\end{figure*}

\begin{table*}
\begin{center}
\begin{tabular}{llcccc}
\hline
& & Vehicle & Bicycle & Chair &  Motor \\
\hline\hline
1  &   \# Models & 1452  & 1204 & 1041 & 1325 \\
2  &   \# Train/ \#Test & 1260/192  & 1082/122 & 925/116 & 1219/106  \\
3  &   \# Semantic labels & 9  & 5 & 7 & 6   \\
4  &   \# Templates & 89640  & 5500 & 15960 & 15600   \\
\hline
\hline
5 &  Basline(SAE-Net+NMS)  & 60.7  & 79.0 & 82.9 & 75.4 \\
6 &  Song et al.~\cite{Song2017} & 66.4  & 84.7 & 73.1 & 84.5   \\
\hline
\hline
7 &    Ours(w/o Semantic template) & 63.1  & 76.4 & 84.4 & 71.3 \\
8 &    Ours(w/o confidence score)  & 67.2  & 84.3 & 88.5 & 83.6 \\
9 &    Ours(SAE-Net+SAI)  & \textbf{72.5} & \textbf{91.0} & \textbf{92.4} & \textbf{87.3}   \\
\hline
\end{tabular}
\end{center}
\caption{Accuracy of semantic abstraction (average Intersection of Union, in percentage) with our benchmark dataset.
Row 1: The number of 3D shapes in our dataset.
Row 2: Training / testing split (number of models) of our dataset.
Row 3: The number of annotated semantic labels for each category.
Row 4: The number of template boxes for each category.
Row 5-9:Average IoU of baseline method, state-of-the-art method, and our method in different settings.}
\label{table:headings}
\end{table*}

\subsection{Benchmark Dataset}

To facilitate quantitative evaluation, we provide an annotated semantic abstraction benchmark dataset.
The 3D shapes in our dataset are collected from ShapeNet~\cite{ShapeNet2015} and 3D warehouse~\cite{Tri3Dwarehouse}. We manually annotate each 3D shape in our dataset using our interactive annotation tool. Details about the annotation tool are provided in the supplementary material. The semantic part categories are defined based on WordNet, and summarized with an overview of the benchmark dataset in Figure~\ref{fig:dataset}.

Table~\ref{table:headings} provides detailed statistics about our semantic abstraction benchmark dataset.
For each category, about 80\% shapes are used for training, and the remaining ones for testing (Row 2).
The third row of Table~\ref{table:headings} presents the number of annotated semantic labels for each category.

\subsection{Comparison with the state of the art methods}
\begin{figure*}
\begin{center}
   \includegraphics[width=0.9\linewidth]{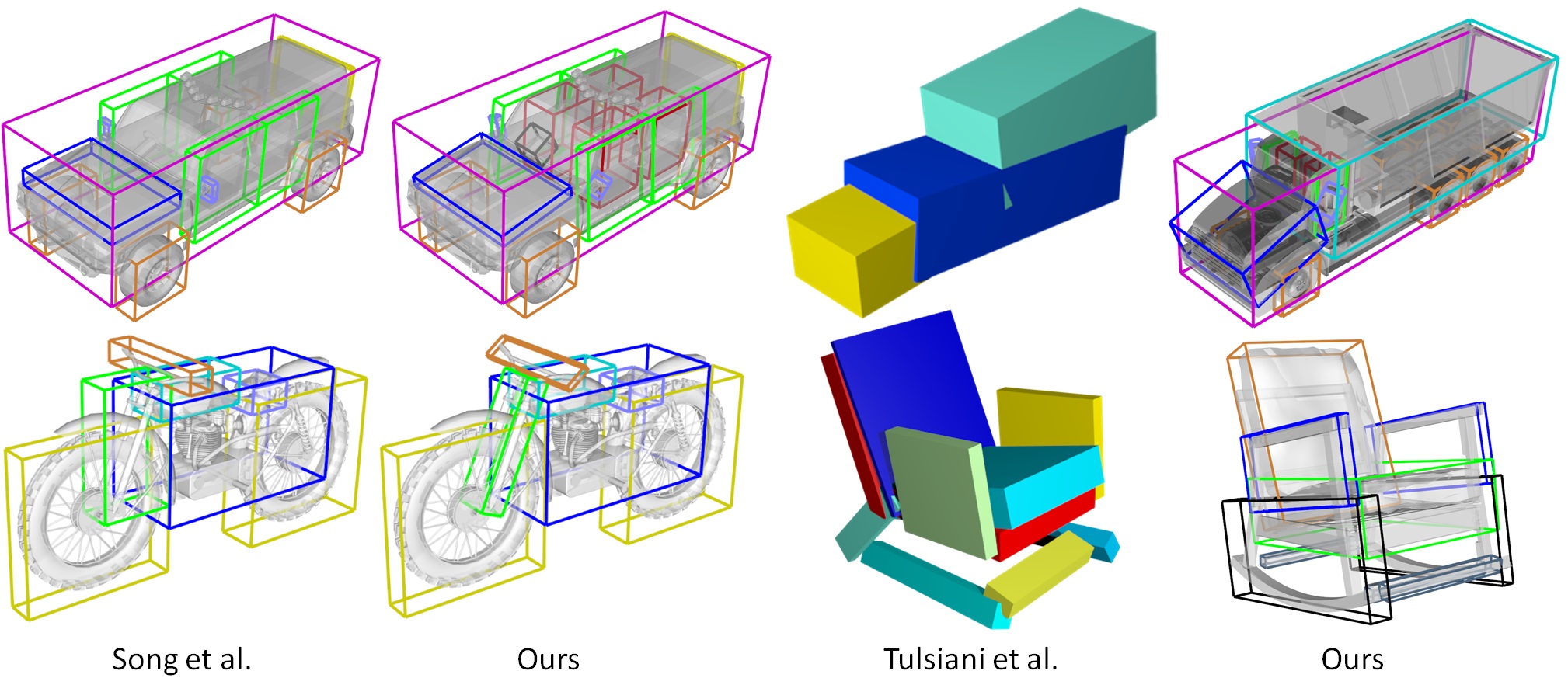}
\end{center}
   \caption{Comparison to Song et al.~\cite{Song2017} and Tulsiani et al.~\cite{abst_17}.}
\label{fig:cmp_with_art}
\end{figure*}

We compare our approach with Song et al.'s method~\cite{Song2017} in our benchmark dataset.
This~\cite{Song2017} is a multi-view based method, which has good effect for the 3D surface model.
However, 
the part beneath the surface of the 3D object cannot be abstracted, as it is not visible.
As shown in Figure~\ref{fig:cmp_with_art},
the seats and steering wheel in the car are completely occluded by the car shell and thus cannot be abstracted using this method~\cite{Song2017}.
Moreover, this method~\cite{Song2017} only estimates the Axis Align Bounding Box (AABB) of the 3D shape's semantic parts,
while our method enables more compact semantic abstraction.
We also report per-category IoU performance of the method~\cite{Song2017} on our benchmark dataset; see Table~\ref{table:headings}(Row 6).
The result indicates a significant  advantage of our semantic abstraction method.
In particular, the improvement in average IoU~\cite{Song2017} ranges from 6.1\% to 19.3\%.

Our method is also compared with Tulsiani et al.'s~\cite{abst_17} method, a state-of-the-art unsupervised deep learning based method for 3D shape abstraction. 
To ensure good performance of the method on our benchmark dataset, we use their model pre-trained on ShapeNet and fine-tune it to our training dataset.
As their method is unsupervised, it is naturally unable to analyze a semantic abstraction. Moreover, as shown in Figure~\ref{fig:cmp_with_art}, we can find that, for complex 3D shapes with overlap, the unsupervised loss function only aims for the maximum coverage of the 3D shape by the smallest possible abstract primitive. 

\subsection{Parameter analyses and ablation studies}

\noindent
\textbf{Evaluation on semantic abstraction candidates generation.}
To verify the effectiveness of our semantic abstraction candidates, we design an intuitive baseline method to generate semantic abstraction candidates. 
For a given 3D shape, we first convert it to the volumetric representation as an occupancy grid with resolution 32 *32*32. For each occupancy voxel location, we will predict $N$ semantic candidates. Each of the proposals corresponds to one of the $N$ boxes with various sizes. In our case, based on statistics of semantic parts sizes in our dataset, we define a set of $N$=20 boxes. The experimental result is reported in row 7 of Table~\ref{table:headings}. For all categories, our method achieves much higher performance when incorporating our generation algorithm. The main reason is that the significant position and scale variation of different semantic parts makes it difficult for uniform sampling.

\noindent
\textbf{Semantic abstraction performance without confidence score.}
For each semantic abstraction candidate, a confidence score is output by our network (Section~\ref{SAE-Net}), which measures how likely it represents a reasonable semantic abstraction. This score is employed in defining the weight in the composite inference (Section~\ref{Composite}). To test its effect, we experiment an ablated version of our method without considering this confidence score (by setting $s(g)$ = 1 in Equation~\ref{eq:CRF})), while keeping all other parameters unchanged. The experimental results are reported in row 8 of Table~\ref{table:headings}. For all categories, our method works better when incorporating confidence score. In particular, the improvement of average IoU over 'w/o confidence score' ranges from 3.7\% to 6.7\%.

\begin{figure}
\begin{center}
   \includegraphics[width=0.9\linewidth]{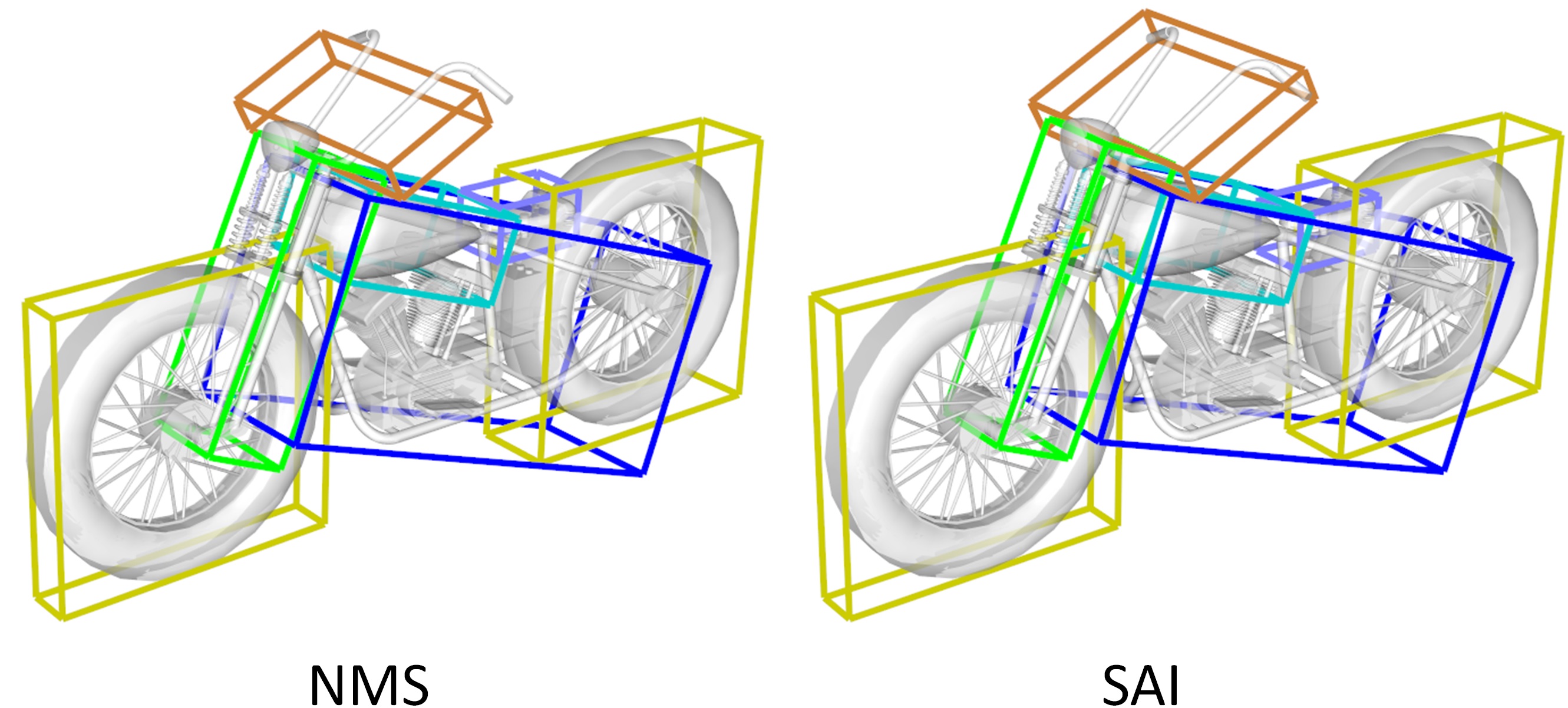}
\end{center}
   \caption{Semantic abstract by NMS (left panel) vs SAI (right panel).}
\label{fig:cmp_with_NMS}
\end{figure}

\noindent
\textbf{Comparison to alternative methods (Non-Maximum Suppression based semantic abstraction estimation).}
To demonstrate the validity of our semantic abstract integration (SAI) method, we also directly merge these predicted results by applying Non-Maximum Suppression (NMS) to generate the semantic abstraction.
In Table~\ref{table:headings}(Row 5 and 9), we report the merging performance (average IoU) over different categories.
We find that SAI performs better. 
This is because NMS simply selects the best among predicted results, which does not ensure coverage of all corresponding semantic parts. 
As shown in Figure~\ref{fig:cmp_with_NMS}(left), there is still a large deviation between the optimal handlebar abstraction selected by NMS and the handlebar itself. However, our method (Figure~\ref{fig:cmp_with_NMS}(right)) takes into account all the predicted results and obtains a better semantic abstraction.

\begin{figure}
\begin{center}
   \includegraphics[width=0.99\linewidth]{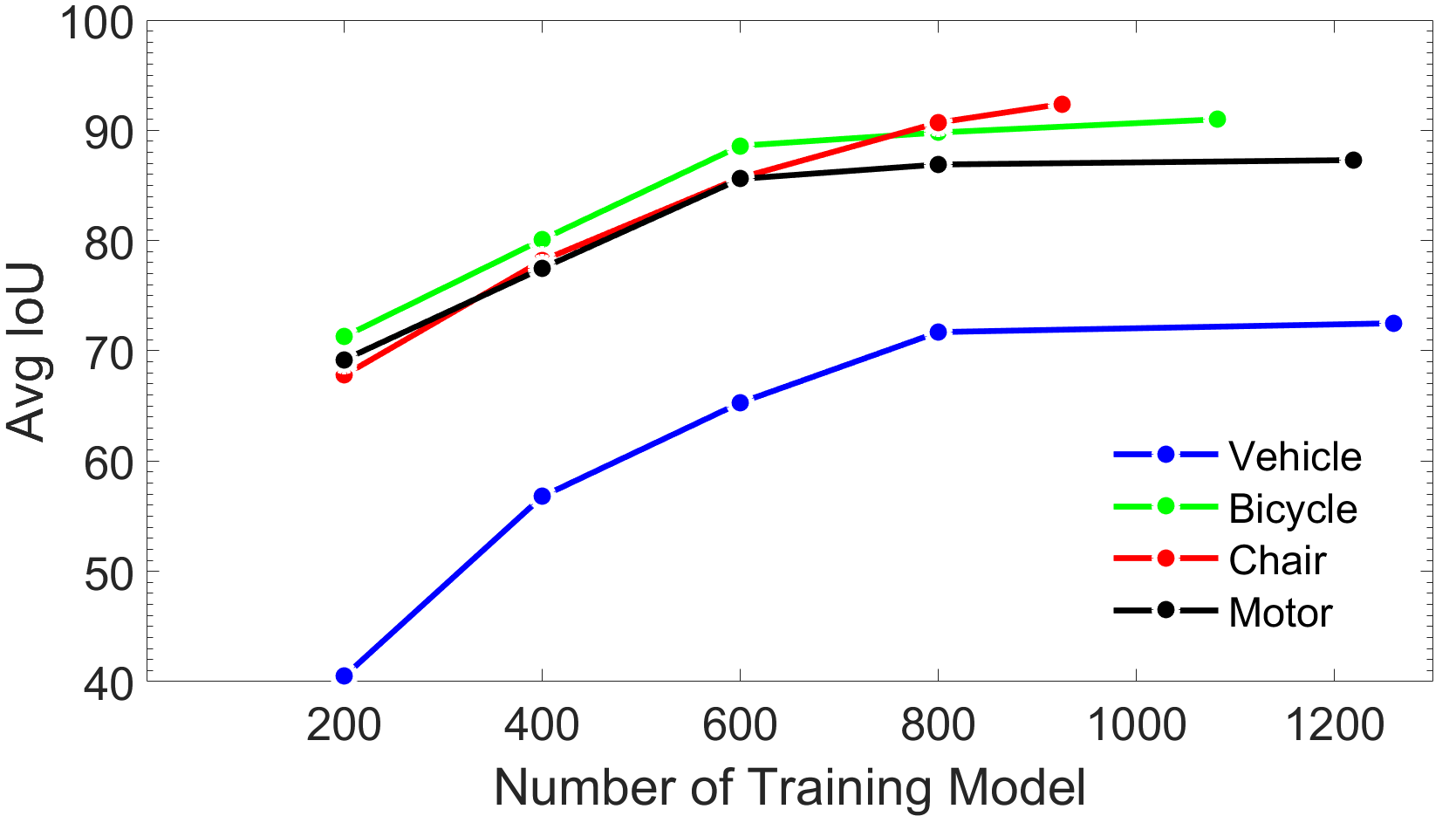}
\end{center}
   \caption{Semantic abstract by NMS (left panel) vs SAI (right panel).}
\label{fig:IoU_vs_Count}
\end{figure}

\noindent
\textbf{Effect of number of training data over Performance.}
Table~\ref{table:headings}(Row 9) shows that our method performs comparably at all training data. 
Here we study the effect of different number of training data for our network. 
Figure~\ref{fig:IoU_vs_Count} shows the plots of average IoU over different number of training data (the total number of training data for each category is shown in row 2 of Table~\ref{table:headings}). The observation is that the performance grows as the number of training data increases, but stops growing at a specific number. For all categories, we found that the number is around 600 to 1000, even for structurally complicated categories such as vehicle and bicycle.

\subsection{Instance-level semantic segmentation}
First we determine the semantic abstraction to which each face belongs.
If the current face belongs to only one semantic abstraction, we can determine its semantic category.
If the current face belongs to multiple semantic abstractions, the current face belongs to the boundary region.
Therefore, we conduct probability voting through the semantic of local neighbors to determine the semantic category.

\begin{figure*}
\begin{center}
   \includegraphics[width=0.9\linewidth]{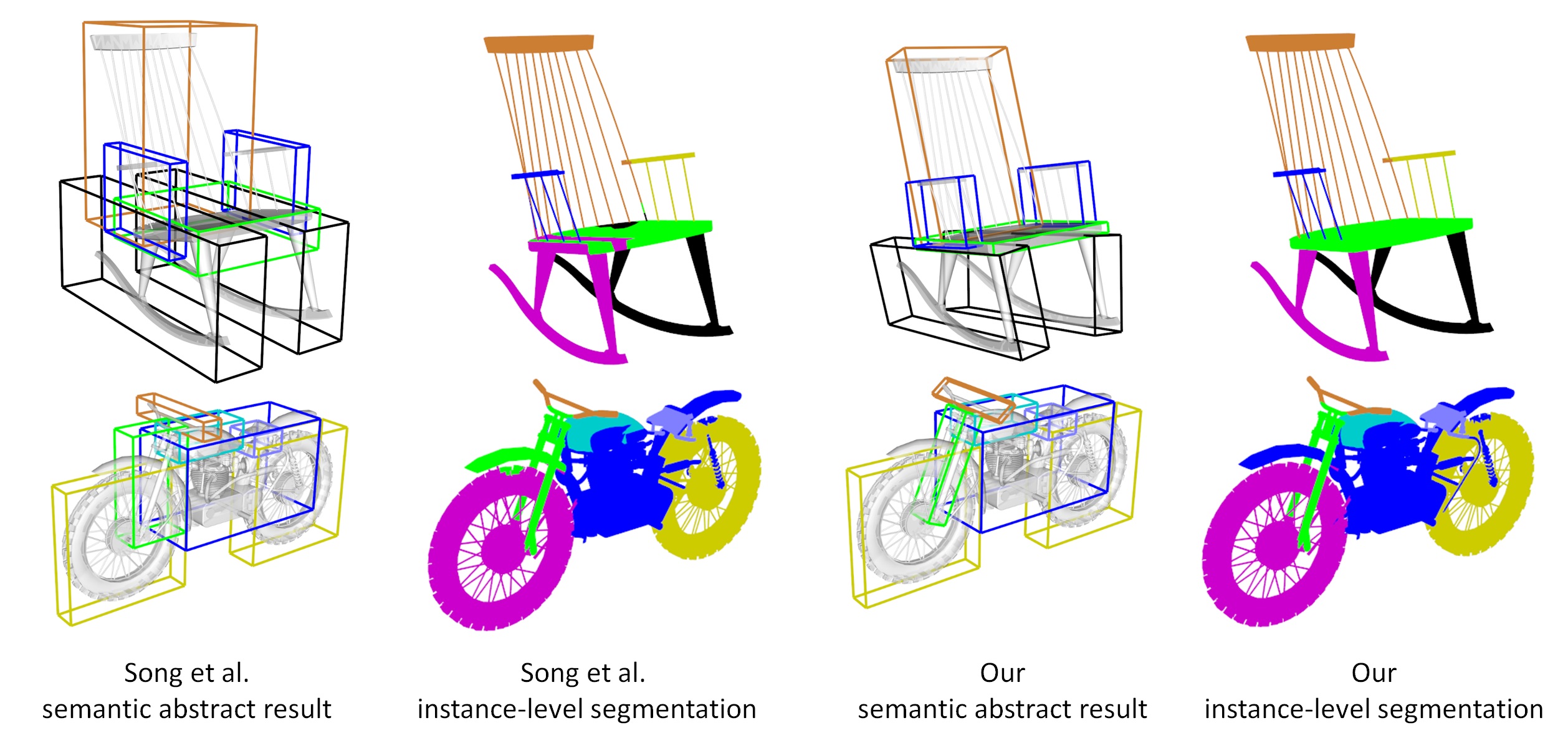}
\end{center}
   \caption{Comparison with Song et al.~\cite{Song2017}. Two examples of 3D shape instance-level semantic segmentation.}
\label{fig:instance_seg}
\end{figure*}

Specifically, let $F$ be the set of faces that already have an identified semantic category. 
For a face $f$ belongs to different semantic abstractions, we first select $K$ nearest neighbors from the set $F$ based on the connectivity between the faces. 
We then determine the semantic category of the current face $f$ by voting, using semantic information on K nearest neighbors.
Figure~\ref{fig:instance_seg} shows a comparison between our method and Song et al.'s method~\cite{Song2017} for instance-level semantic segmentation.

Note that, the 3D shapes in our dataset are collected from ShapeNet~\cite{ShapeNet2015} and 3D warehouse~\cite{Tri3Dwarehouse}. 
As such, they are likely assembled by many components rather than manifold models. Hence, we cannot directly use graph cut optimization with these shapes.

\begin{figure}
\begin{center}
   \includegraphics[width=0.9\linewidth]{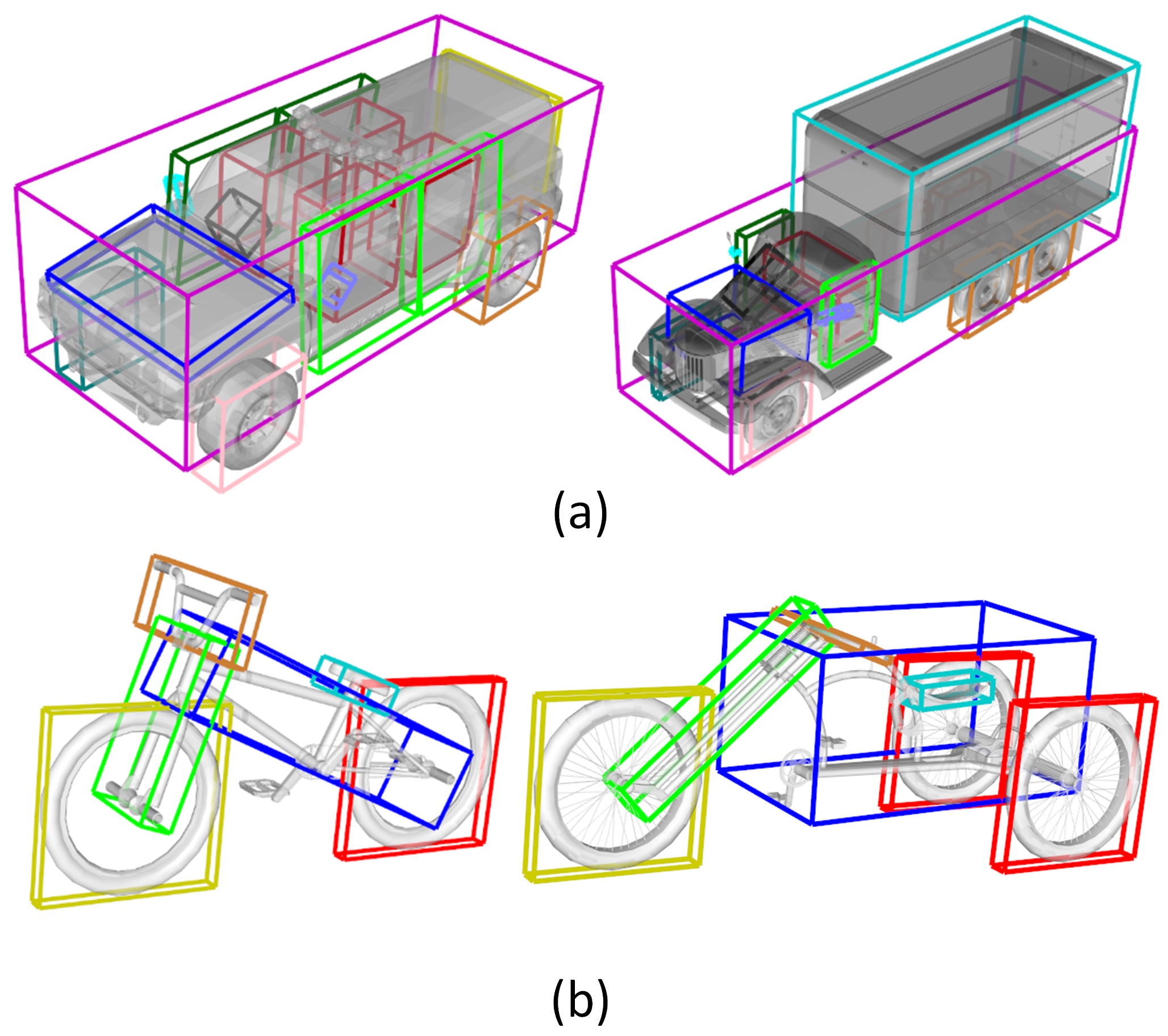}
\end{center}
   \caption{Two examples of 3D shape semantic matching.}
\label{fig:shape_match}
\end{figure}

\subsection{Shape semantic matching}
We also apply our approach to shape semantic matching . Specifically, for a pair of shapes, we first analyze the semantic abstraction, which would facilitate the global alignment of the two shapes based on  this semantic information. In the correspondence search, we assume that two corresponding abstractions have the same semantics, so we can greatly reduce the search space using the constraint conditions.
According to the experimental results (Figure~\ref{fig:shape_match}), this hypothesis is reasonable, and especially suitable for the models with complex topology. Specifically, for a pair of source group and target group, 
in which these abstracts have the same semantics, we first normalize these two groups, and then perform a nearest neighbor search for the corresponding abstractions within the source group. 
Meanwhile, to ensure the integrity of the search results, we conduct a two-way search. However, this may produce a large number of repeat matches, and to address this, we remove the same matches. Finally, for many-to-many matching, we aggregate the search results with the same abstraction. 
Figure~\ref{fig:shape_match} shows two examples of shape semantic matching. Note that our approach does not introduce any high-level properties, such as symmetry.

\section{Conclusions}
This paper introduces a novel approach for joint estimation of 3D shape abstractions and semantic analysis. We first generate a number of 3D semantic abstraction candidates for a 3D shape. We then employ these candidates to directly predict the semantic categories and refine the parameters of the candidate regions simultaneously using a deep convolutional neural network. Finally, we design an algorithm to fuse the predicted results and obtain the semantic abstraction, which is shown to perform better than standard non- maximum suppression.
Experimental results demonstrate that our approach can produce
state-of-the-art results. Moreover, we also show that our results can be
easily applied to instance-level semantic part segmentation and shape
matching.

\noindent
\textbf{Limitations.}
The main limitation of our method is that our method can only train each type of model independently.
Meanwhile, our method requires some preprocessing and postprecessing.
As a future work, we would consider incorporating all steps into the network to make the entire model end-to-end trainable.

\noindent
\textbf{Acknowledgements.} We thank the anonymous reviewers for their valuable comments. 
This work was supported by National Natural Science Foundation of China (U1736217).

{
\bibliographystyle{eg-alpha-doi}
\bibliography{cvmbib}
}


\end{document}